\pdfoutput=1

\documentclass[10pt,twocolumn,letterpaper]{article}

\usepackage[pagenumbers]{cvpr} 

\usepackage{graphicx}
\usepackage{amsmath}
\usepackage{amssymb}
\usepackage{booktabs}

\usepackage[pagebackref,breaklinks,colorlinks]{hyperref}

\usepackage[capitalize]{cleveref}
\crefname{section}{Sec.}{Secs.}
\Crefname{section}{Section}{Sections}
\Crefname{table}{Table}{Tables}
\crefname{table}{Tab.}{Tabs.}

\def\cvprPaperID{4323} 
\def\confName{CVPR}
\def\confYear{2022}

\begin{document}

\title{Local Relighting of Real Scenes}

\author{Audrey Cui\\
MIT\\
{\tt\small audcui@mit.edu}
\and
Ali Jahanian\\
MIT\\
{\tt\small jahanian@mit.edu}
\and 
Agata Lapedriza\\
Universitat Oberta de Catalunya\\
{\tt\small alapedriza@uoc.edu}
\and
Antonio Torralba\\
MIT\\
{\tt\small torralba@mit.edu}
\and
Shahin Mahdizadehaghdam  \\
Facebook\thanks{Shahin Mahdizadehaghdam and Rohit Kumar were at Signify when this work was done.}\\
{\tt\small shahinaghdam@fb.com}
\and
Rohit Kumar \\
Sage\footnotemark[1]\\ 
{\tt\small rohit.kumar2@sage.com}
\and
David Bau \\
Northeastern University \\
{\tt\small davidbau@northeastern.edu}
}
\let\oldtwocolumn\twocolumn
\renewcommand\twocolumn[1][]{%
    \oldtwocolumn[{#1}{

\begin{center}%
\centering%
\captionsetup{type=figure} 
\includegraphics[width=\textwidth,trim=0 0 0 0]{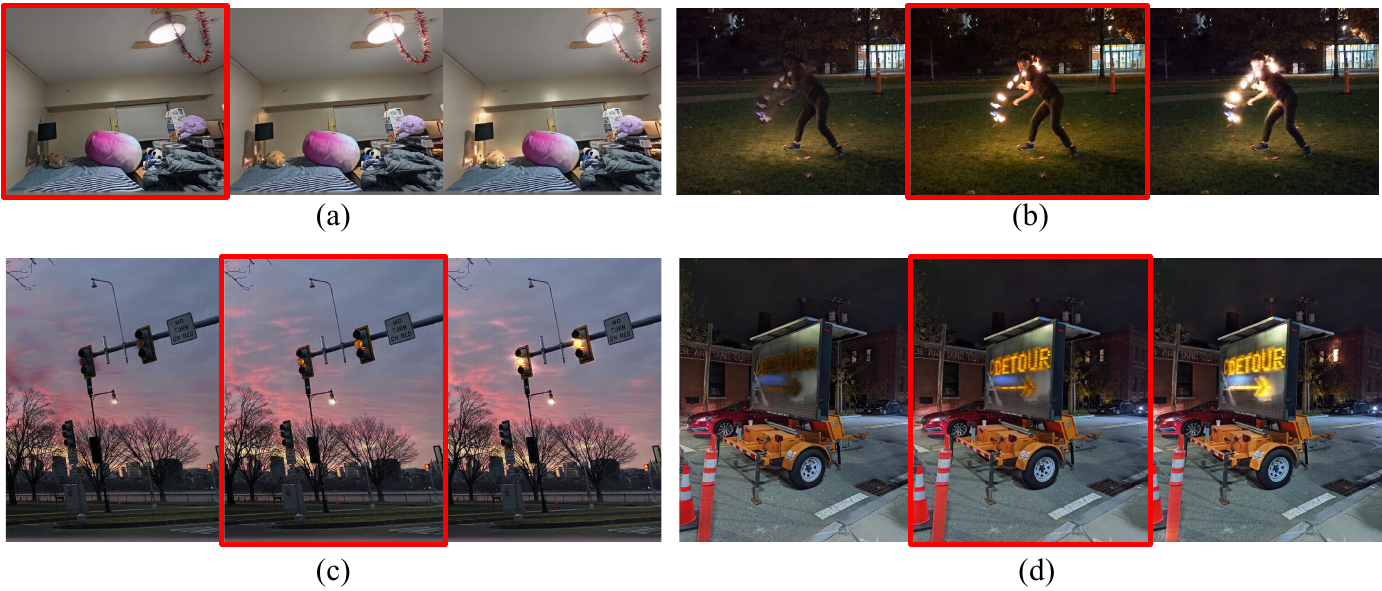}\\
\captionof{figure}{Can we use deep generative models to edit  real scenes ``in the wild"? We propose using a pretrained generator model to generate an unbounded dataset of paired images to train an image-to-image translation model to relight visible light sources in real scenes. In the examples above, the original photo is highlighted with a red border. We show that our unsupervised method is able to turn on light sources present in the training domain (i.e. lamps) in (a). Additionally, our method can detect and adjust prelit light sources that are way outside its training domain, such as fire fans (b), traffic lights (c), and road signs (d). }
\label{fig:teaser}
\end{center}
    }]
}

\maketitle

\begin{abstract}
   We introduce the task of local relighting, which changes a photograph of a scene by switching on and off the light sources that are visible within the image. This new task differs from the traditional image relighting problem, as it introduces the challenge of detecting light sources and inferring the pattern of light that emanates from them. We propose an approach for local relighting that trains a model without supervision of any novel image dataset by using synthetically generated image pairs from another model.  Concretely, we collect paired training images from a stylespace-manipulated GAN; then we use these images to train a conditional image-to-image model. To benchmark local relighting, we introduce Lonoff, a collection of 306 precisely aligned images taken in indoor spaces with different combinations of lights switched on. We show that our method significantly outperforms baseline methods based on GAN inversion. Finally, we demonstrate extensions of our method that control different light sources separately. We invite the community to tackle this new task of local relighting. 
\end{abstract}


\section{Introduction}
\label{sec:intro}

The desire to turn on or off the lights in a photograph is a longstanding challenge in computer vision: changing lighting is widely useful, but it is also difficult because the physics of light transport make no distinction between a photon that arrives due to a bright intrinsic color, a well-aligned surface normal, or a strong light source.  The problem of altering a 2d scene to change the direction of distant ``global light'' that illuminates the objects in a scene is a topic of ongoing research~\cite{el2021ntire}.  However, the task of altering what we call \emph{local relighting}, in which the light source to be changed is itself within the scene, has not yet been characterized.




Our paper poses the following questions: Is it possible to relight a 2d image where the light source is visible in the scene? If so, how can it be done, and how can success be measured?

\begin{figure*}[t]
  \centering
    \includegraphics[width=\textwidth]{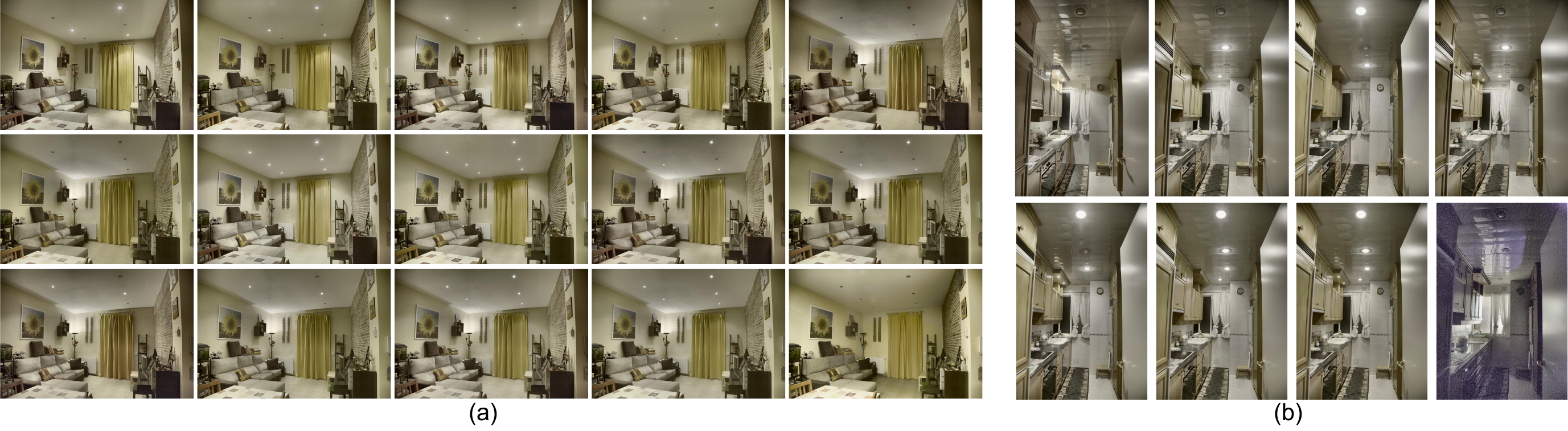}
  \caption{
Two examples of spaces included in the Lonoff Dataset: a dining room (a) and a kitchen (b). Per each of these two spaces we show the images corresponding to all the different lighting combinations that we can have in each of the spaces. 
  }
  \label{fig:lonoff_sample}
\end{figure*}

The first significant problem is the absence of any existing dataset that establishes a ground truth that shows exactly how a scene would appear if local lights are changed.  Therefore, we introduce a new dataset of ground-truth images of identical scenes under different local lighting conditions.  The dataset contains 306 images of 9 scene categories in which a set of photographs is precisely aligned, in which the only change is the the lighting condition. Figure~\ref{fig:lonoff_sample} illustrates some examples of this curated dataset.  The cost of collecting such a dataset is high, so the scale of the dataset is useful for benchmarking, but not training a large model.

The second significant problem is that presence of the light source in a scene introduces the challenge of recognizing the light source as well as the illumination attributed to it. Classic computer graphics techniques are insufficient, so we turn to deep learning.  Typically deep learning methods require a wealth of labeled training data in order to solve a problem, but in this case we have only a small number of paired images.  We shall demonstrate that nevertheless, it is possible to design a nearly fully unsupervised training procedure that creates a model that can relight many local light sources in real scenes.  The training relies only on unpaired images; it can be done without image labels nor collection of a large dedicated lighting dataset. As part of our contribution, we show how to use a pretrained image synthesis model to generate paired training images, giving us an unbounded number of training samples.
This data source is used to train an image-to-image translation model to apply the relighting technique to real images ``in the wild'' as illustrated in Fig.~\ref{fig:teaser}.

Furthermore, we demonstrate that by exploiting a widely-available object segmentation model, we can extend the training method to provide fine-grained control that allows a user to select which light to turn on or off.

In this paper we make the following contributions:
\begin{itemize}
	\item We introduce a carefully curated dataset of scenes under varied lighting conditions as a ground truth for benchmarking the local relighting challenge.
	\item We introduce an unsupervised method for tackling local relighting by using a pretrained model to generate an unbounded number of paired data as the only data source to train another model to relight real scenes.
	\item We introduce a ``user selective" method that allows users to control which lights are to be turned on and off. Our method exploits  segmentation information during the training process,.
\end{itemize}
Our code can be accessed  \href{https://github.com/audreycui/relight}{here}. Our benchmark dataset will be made available upon publication. 


\section{Related Work}
\vspace{3pt} \noindent \textbf{Image Relighting.}
Image relighting is the problem of rendering a scene under different and novel lighting conditions. Traditional methods addressed this problem by modelling the light transportation function, i.e., the function that maps the incident illumination from a specific direction to the radiance at each pixel of the image. This modelling was done by gathering a large number of images from the same scene in order to interpolate new lighting conditions \cite{mahajan2007theory, sloan2003clustered, kalantari2016learning, malzbender2001polynomial}. In contrast to the need of having large collections of images of the same space to model the light transportation function, Xu et al. \cite{xu2018deep} presented a deep learning  approach based that only requires 5 images of a scene to reproduce scene appearance under any directional light lying in the visible hemisphere. They accomplish this training a deep convolutional neural network for relighting using a large synthetic dataset. More recently, El Helou et al. \cite{helou2020vidit} introduced the VIDIT dataset, a collection of images showing versions of the same scene under varying global lighting conditions, with the light source being out of frame, facilitating the training of deep learning models. Later, El Helou et al. \cite{el2021ntire} has posed the task of reconstructing a given scene with new global lighting settings of an input image, given the scene's depth map. The depth map enables approaches that construct a physical understanding of the scene, such as using features for texture and structure \cite{yang2021multi}, for albedo and shading \cite{yazdani2021physically}, or disentangling intrinsic structure from lighting \cite{wang2021multi}. Using VIDIT, \cite{gafton20202d} trained a conditional image to image translation model to relight images in the 8 cardinal directions. For the problem of local relighting, there is no existing dataset depicting the scenes under varying local lighting that is large enough to reliably train a complex model.

\vspace{3pt} \noindent \textbf{Semantic image editing.}
StyleGAN2 \cite{karras2019style} \cite{karras2020analyzing}, a state of the art image synthesis model, maps a latent $z$ into another latent space $w$ to create a representation disentangled for meaningful image attributes. Realistic semantic image edits can be made by steering latents in optimized directions \cite{jahanian2019steerability, shen2020interpreting, harkonen2020ganspace}, or by finding and activating neurons that encode  semantic concepts \cite{bau2018gan}. Bau et al. \cite{bau2021paint} used a user specified mask to decouple $w$ latents, enabling region specific edits while preserving the content of unmasked areas. Wu et al. \cite{wu2021stylespace} discovered that the channel-wise style latent space (StyleSpace) of StyleGAN is significantly more disentangled than other latent spaces. Thus, manipulation of StyleSpace enables fine control of specific image attributes. 

\vspace{3pt} \noindent \textbf{GAN inversion methods.}
The semantic editing methods described above can only be directly applied onto generated images. In order to apply such semantic edits to real images, the image must be inverted into a GAN latent space representation first, which is a nontrivial challenge. Existing GAN inversion methods include \cite{tov2021designing, karras2020training, gu2020image, zhu2020domain}. 
Our method works around the difficult GAN inversion problem by using image-to-image translation methods instead to preserve the structure of original image, similar to \cite{Viazovetskyi2020StyleGAN2DF}. However, while \cite{Viazovetskyi2020StyleGAN2DF} requires a  classifier pretrained with facial attribute labels to train their image-to-image translation model, our training method does not require supervision.

\section{Lonoff Dataset}

The "Lights on/off" Dataset (Lonoff) is a collection of images taken in indoor spaces under different illumination conditions. Concretely, in each space we collected several different pictures, including different combinations of light sources switched on. In general, when the switches allow to switch on each light separately, all the possible light combinations are included for each scene. For example, if a space contains three light sources or lamps, denoted by $l_1, l_2, l_3$, the dataset contains a total of $7$ images of this space: $3$ images with just one light switched on, $ \binom{3}{2}$ images with two lights switched on ($\{l_1,l_2\}$, $\{l_1,_3\}$,  $\{l_2,l_3\}$), and 1 image with all the three lights switched on. The images are taken using a tripod so that the location of the camera is the same for all the images collected in the same space. Fig.\ref{fig:lonoff_sample} shows all the images of two of the spaces included in the Lonoff dataset: a dining room (a) and a kitchen (b). Notice that in the example of the kitchen, the light sources include lamps as well as a window. In particular, the last image of the kitchen space shows the kitchen illuminated just with the light that comes through the widow, with all the lamps switched off.

\begin{figure}
  \centering
    \includegraphics[width=\columnwidth]{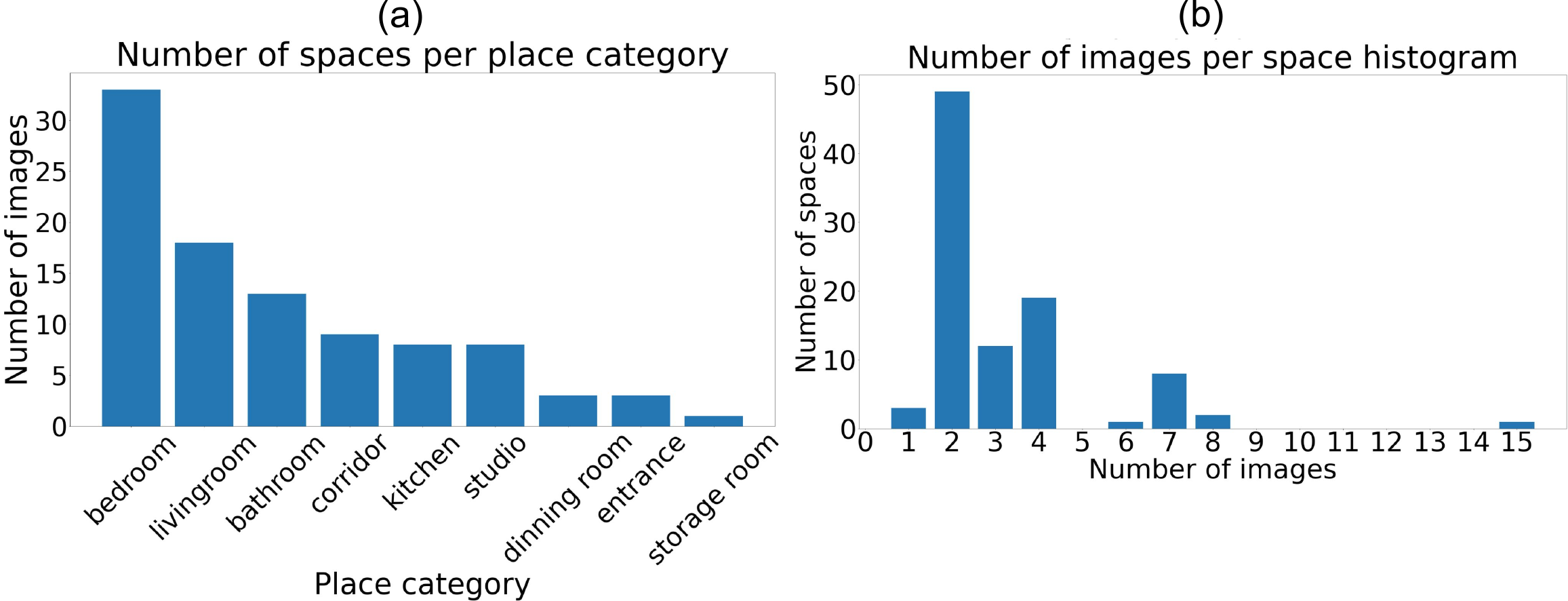}
  \caption{
Number of spaces included in the Lonoff Dataset per place category (a) and number of images per space histogram (b). 
  }
  \label{fig:lonoff_stats}
\end{figure}

The Lonoff dataset contains images of $9$ place categories: bathroom, bedroom, corridor, dinning room, entrance, kitchen, living room, storage room, and studio. Fig.\ref{fig:lonoff_stats}.a shows the number of spaces per place category and Fig.\ref{fig:lonoff_stats}.b shows the number of images per space histogram. We observe there is just one space that contains 15 different pictures, with is the living room shown in Fig.\ref{fig:lonoff_sample}.a. 

The dataset also includes the following manual annotations: segmentation of all the light sources, light source category (ceiling recessed light, floor lamp, fluorescent tube, flush mount light, light troffer, pendant lamp, sconce, spotlight, table lamp, or window), segmentation of the light source parts (e.g. column, shade, arm, aperture, backplate), and the on/off attribute per light source, indicating whether the light source is switched on or not.  

The intent of the Lonoff dataset is to provide a detailed and curated dataset for testing indoor illumination understanding models.
\begin{figure*}
  \centering
    \includegraphics[width=\textwidth]{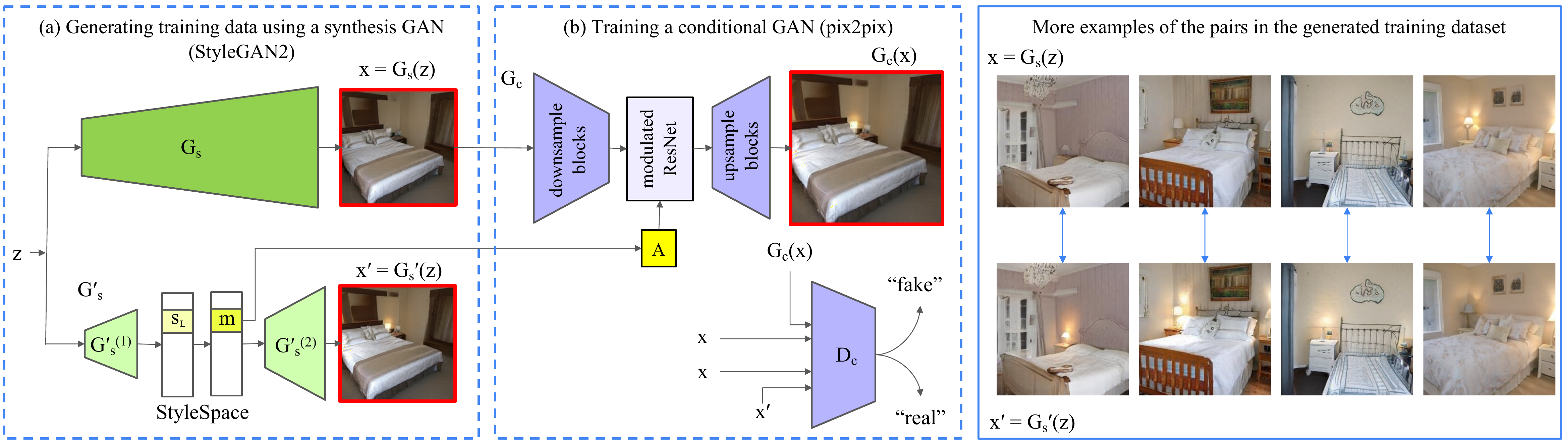}
  \caption{
  An overview of our unsupervised method for relighting scenes. (a) We use $G_s$ to generate paired scenes for training data, $x$ and $x^{\prime}$. $x$ is synthesized from a random $z$ without modifying $G_s$. We modify $G_s$ into $G_s^{\prime}$ by setting its StyleSpace channel $s_L$ to a scalar $m$. $x'$ is synthesized with the same $z$. (b) To train $G_c$ to relight images, we pass $x$ as the training input to generate $G_c(x)$. An affine transform $A$ is applied on $m$ to modulate the ResNet bottleneck of $G_c$, teaching $G_c$ to relight scenes to varying intensities. $G_c(x)$, the ``fake" image, and $x^{\prime}$, the ``real" image are both conditioned on $x$ by the discriminator to obtain a discriminative loss. Note that while $x^{\prime}$ is technically a fake image in that it's generated, we denote it the ``real" image because it serves as the image-to-image translation ground truth.}
  \label{fig:architecture}
\end{figure*}
\section{Methods}
In this section, we formalize our objectives on how to train a generative model by learning from another generative model as the source of data, how to relight scenes without supervision, and how to selectively edit light sources.

\subsection{Using a GAN to generate training data}

How do we train a model to relight a visible light source without a large training dataset? An ideal training dataset would supply a large number of paired examples of scenes in which the only difference between the pair of images is a change in lighting. However, collecting such a large-scale dataset is prohibitive.

Instead, we propose using a pretrained generative model as a source of training data.  It has been observed that state-of-the art GANs such as StyleGAN2 disentangle meaningful factors of variation in their latent channels.~\cite{wu2021stylespace}  For example, by altering a single channel, StyleGAN2 can change a single image attribute, such as a person's hairstyle, or a car's wheel angle.  We have observed that StyleGAN2 will also disentangle a channel that selectively controls the lighting in a scene.

Therefore we use a GAN to generate pairs of training images: each pair of images is generated using the same latent code, with the only difference being the value of the lighting-control channel. Fig.~\ref{fig:architecture} shows an example of the training data that can be generated.



Formally, we use a StyleGAN generator $G_s:z \rightarrow x$ as the source of data where we obtain a datapoint $x$ from a random vector $z$.  In practice, we use a pretrained model trained to generate bedroom images. We modify this generor and call it $G^{\prime}_s$ as follows: We divide this generator into two subnetworks $G^{\prime(1)}_s$ and $G^{\prime(2)}_s$ in two steps
\begin{align}
    s & = G^{\prime(1)}_s(z) \\ 
    x^{\prime} & = G^{\prime(2)}_s(s) \label{eq:modified} 
\end{align}
where the style vector $s$ contains disentangled components.  
Denote the lighting component $s_L$ as the channel that gives the lighting effect.


To identify a lighting channel $s_L$, we directly annotate some of the pixels corresponding to lighting in a single image and then select the channel that influences those pixels the most, a method inspired by \cite{wu2021stylespace}. See Appendix \ref{sec:channel_id} for more details.
Because the bedroom images in which our $G_s$ was trained have lamps as the predominant light source, we can identify $s_L$ that specifically control lamp lighting in generated images. 


Given a random $z$, we pass it through $G_s$ twice to generate each training pair. $G_s$ is unmodified during the first pass, giving us the image $x$. For the second pass, we set $s_L$ to a random scalar $m$, which alters the StyleSpace $s_L = m $. 
Applying this to Eq.(\ref{eq:modified}) gives us $x^{\prime}$, which depicts the same scene as $x$ but with the light source modified.   

\subsection{Relighting scenes without supervision} \label{unsupervised}

One approach to relighting might be to invert a scene into into StyleGAN2's latent representation and then regenerate it after manipulating $s_L$.  However, we find that reconstruction of complicated scenes is not accurate after inversion ~(e.g. see Fig.~\ref{fig:grid_agata_dataset}). To circumvent this problem, we formulate relighting as an image-to-image translation problem by training a conditional model $G_c$. Our $G_c$ is based off of pix2pixHD \cite{wang2018high}, which extends from the pix2pix framework \cite{isola2017image}, which is able to accurately reconstruct an input image because of its large internal representations. Although pix2pix does not disentangle a latent for lighting, we can teach it to edit lighting by training it to translate images from.  

Our first task is to relight a scene in an unsupervised manner. Because there is no user input on which specific light source should be relit, this task assumes that all visible light sources should be altered. 

In order to control the intensity of lighting in the relit scene, we introduce modulation to the ResNet blocks \cite{he2016deep} in the bottleneck of $G_c$. Our modulation method is inspired by that of StyleGAN: we apply a trainable affine transform $A$ onto the scalar $m$ to obtain a style vector $A(m)$. 

Let $r$ be the input to each ResNet block, and $F$ be the ResNet block mapping function. Our modulated ResNet block can be represented as: 
\begin{equation}
  r = r + F(r\circ(1 + A(m))),
  \label{eq:mod}
\end{equation}
where $\circ$ denotes the Hadamard product. We add $1$ to $A(m)$ to avoid annihilating $r$ if $A(m)$ is close to $0$. We implement $F$ as a 2 layer convolution block. The bottleneck of our $G_c$ uses $9$ ResNet blocks, the same number as pix2pixHD. 


We keep the discriminator $D_c$ unaltered from pix2pixHD, which calculates the following loss: 
\begin{equation}
  \mathcal{L} = \mathcal{L}_{\text{GAN}} + \lambda \mathcal{L}_{\text{FM}}. \label{eq:totalloss} \\
\end{equation}
We pass the following inputs to the GAN loss $\mathcal{L}_{\text{GAN}}(G_c, D_c)$: 
\begin{equation}
  \mathbb{E}_{(x, x^{\prime})}[\log D_c(x, x^{\prime})] + \mathbb{E}_x[\log(1-D_c(x, G_c(x)))] \label{eq:ganloss_pos} \\
 \end{equation}
and perceptual loss $\mathcal{L}_{\text{FM}}(G_c, D_c)$:
\begin{equation}
 \mathbb{E}_{(x, x^{\prime})} \sum_{i=1}^{T} \frac{1}{N_i} [|| D_c^{(i)}(x, x^{\prime}) - D_c^{(i)}(x, G_c(x)))||_1]
 \label{eq:percloss} \\
\end{equation}
for each layer $i$ in a $T$ layer discriminator with $N_i$ elements in that layer, used by the importance controller $\lambda$. 

\begin{figure}[t]
  \centering
  
    \includegraphics[width=.95\columnwidth]{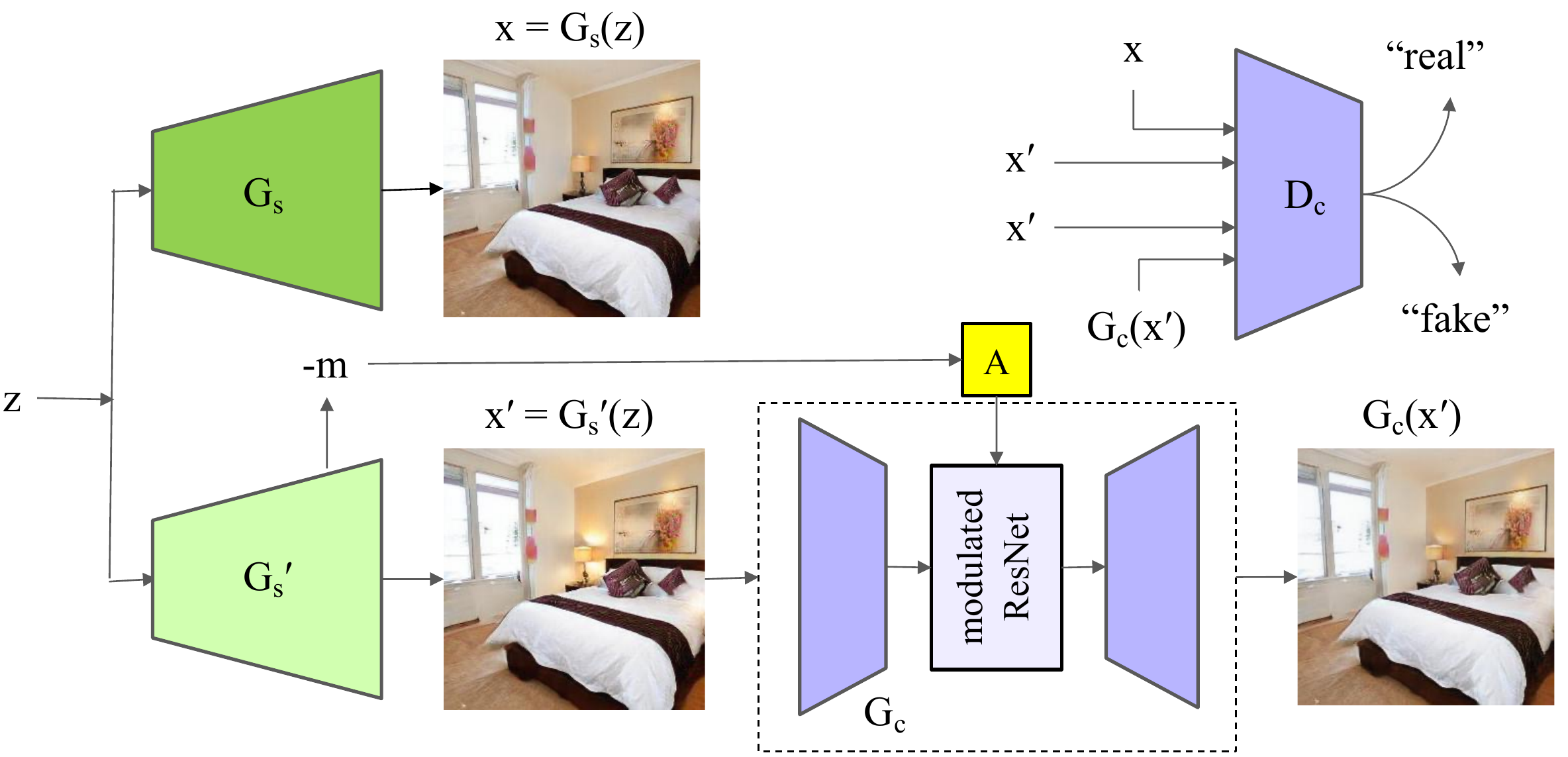}
  \caption{
    Overview of our reverse training method for teaching the model how to turn off lights realistically. We essentially flip the input versus the target for $G_c$. $G'_s(x)$ becomes the target and $G_s(x)$ becomes the input. A negated $m$ modulates the ResNet bottleneck of $G_c$. 
 }
  \label{fig:reverse}
  \vskip\abovecaptionskip 
  \includegraphics[width=0.9\columnwidth]{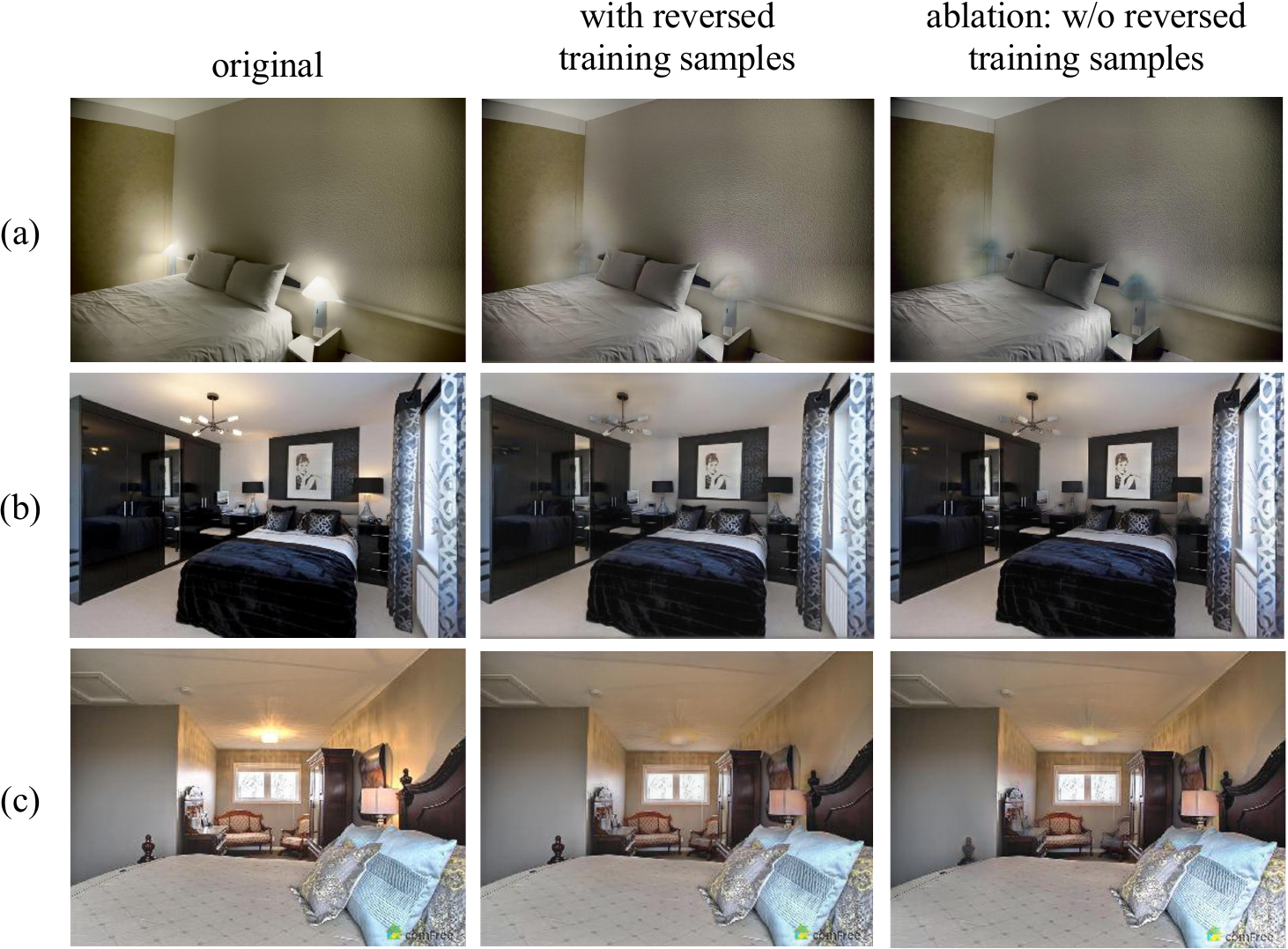}
  \caption{
  Comparing the effect of training with reversed training samples and the ablation of simply using negative $m$ values without reversing training samples. The scalar $m$ is set to the same negative value to generate these examples. In example (a), we see that the ablation creates more unrealistically dark splotches over the lamps compared to our main method. As shown in (b) and (c), the ablation leaves some of the reflected light on the ceiling, whereas our main method removes it more cleanly. 
 }
  \label{fig:lightsoff}
\end{figure}

We notice that while $G_s'$ brightens lights realistically when $m$ is positive, it often creates an unrealistic dark patch over the light source when we set $s_L$ to a \emph{negative} $m$. Because the quality of $G_c$'s relighting results depends on the quality of data it is trained on, we propose using reversed training samples as shown in \cref{fig:reverse} to teach $G_c$ how to turn off lights realistically. 
In that training procedure, we choose a random \emph{positive} $m$, so that $x^{\prime}$ is a brightened version (to varying intensities) of $x$. For half of the training samples, we follow the method illustrated by \cref{fig:architecture}. The other half is ``reversed": we pass the brightened $x^{\prime}$ as the input to $G_c$, modulate ResNet blocks with $-m$, and use $x$ as the target image to calculate discriminative losses $\mathcal{L}_{\text{GAN}}$ and $\mathcal{L}_{\text{FM}}$: 
\begin{align}
  \mathbb{E}_{(x^{\prime}, x)} [\log D_c&(x^{\prime}, x)] + \\ & \mathbb{E}_{x^{\prime}}[\log(1-D_c(x^{\prime}, G_c(x^{\prime})))] \nonumber
\end{align} 
\begin{align}
\mathbb{E}_{(x^{\prime}, x)} \sum_{i=1}^{T} \frac{1}{N_i} [|| D_c^{(i)}(x^{\prime}, x) - D_c^{(i)}(x^{\prime}, G_c(x^{\prime})))||_1]
 \label{eq:percloss_neg}
\end{align}
 
\cref{fig:lightsoff} shows the effect: using reversed training samples allows lamps to be turned off more realistically.

\begin{figure*}
  \centering
    \includegraphics[width=\textwidth]{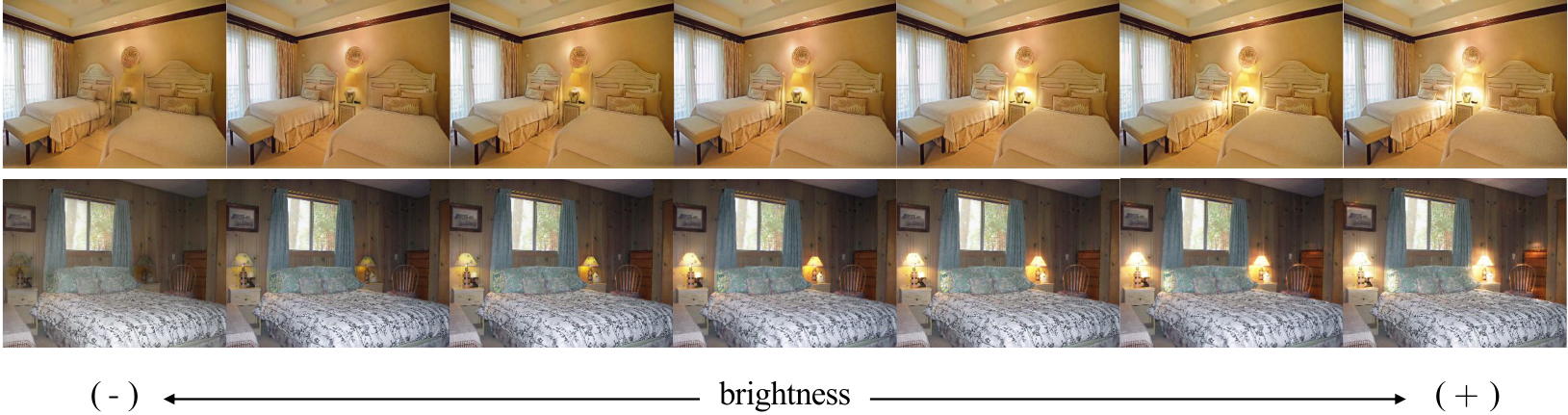}
  \caption{
  Examples are from our unsupervised method. We show that modulating the resnet bottleneck in $G_c$ allows for fine control over relighting intensities in real scenes. This allows us to both brighten and dim light sources to varying degrees.  
  }
  \label{fig:modulation}
\end{figure*}

During inference, $G_c$ takes as input a real photo and a scalar $m$ (which can be either positive or negative), and outputs the photo relit to an intensity $m$. \cref{fig:modulation} shows examples of relighting the same photo to varying intensities. 

\subsection{User selective edits}

\begin{figure}
  \centering
    \includegraphics[width=1\columnwidth]{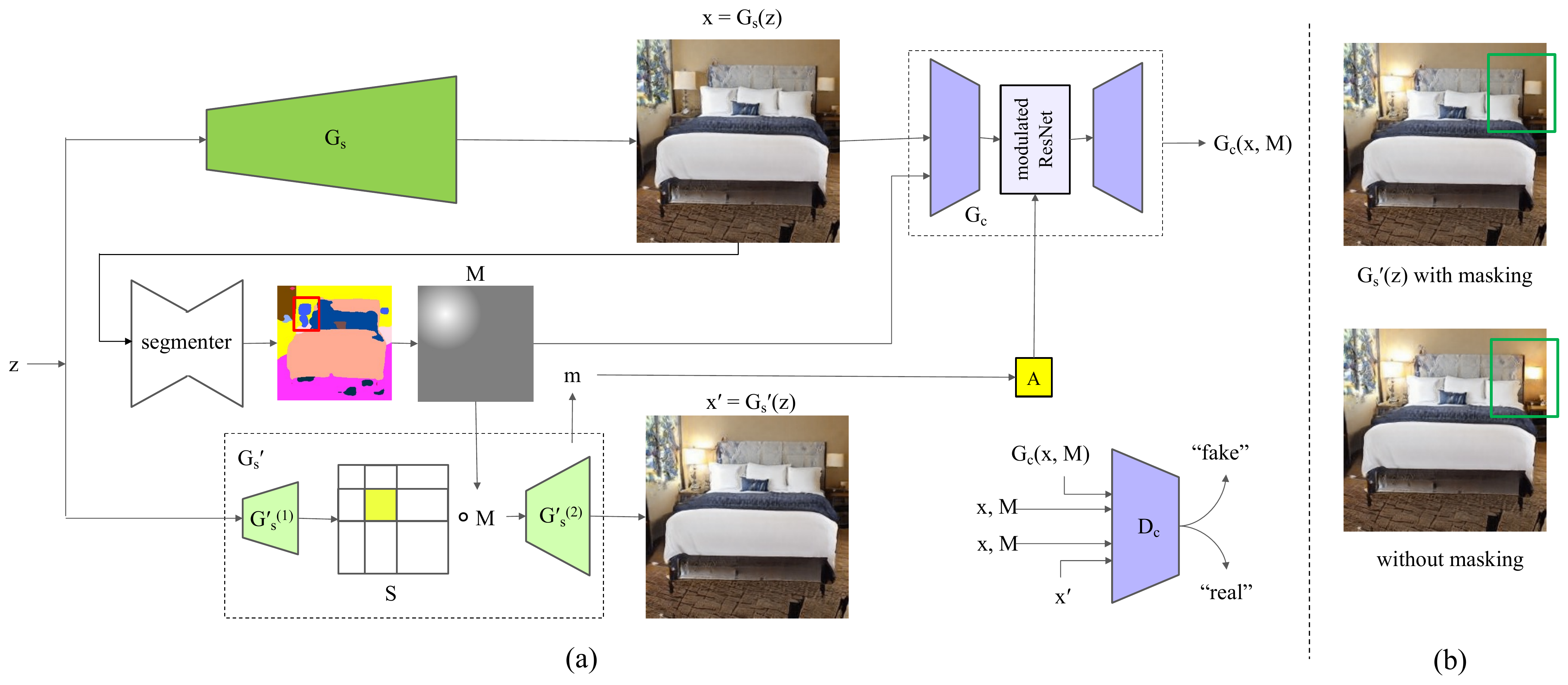}
  \caption{
  (a) shows an overview of our regional editing method. We keep the overall framework of using $G_s$ to synthesize paired training data to train $G_c$. After we generate $x = G_s(z)$, we create a mask $M$ centered at the centroid of the largest lamp present in $G_s$, which we locate using a segmenter. We modify $G_s'$ to only change the lighting channel $s_L$ to a scalar $m$ under the masked region $M$. $G_c$ now takes both $x$ and $M$ as input to generate $G_c(x, M)$. Additionally, $M$ is also conditioned on to calculate discriminative losses. In (b), we show that the masked $S$ successfully brightens just the left lamp without altering the right lamp, whereas the non masked $S$ (approach we take in the unsupervised method) also alters the right lamp.}
  \label{fig:regionaledit}
\end{figure}
Would it be possible to control light sources in the same scene separately? We propose a second task of allowing a user to select which light source to relight. 

As demonstrated in \cref{fig:regionaledit}, we define a light location aware method for modifying $G_s$ into $G_s'$. After $x$ is generated, we use the segmenter model created by \cite{xiao2018unified} pretrained on the ADE20K scene dataset \cite{zhou2017scene} to identify largest lamp present in $x$. We create a mask $M$ centered at the centroid of the largest lamp with radius equal to $1.5$ times the height or width of that lamp, whichever is larger. $M$ fades in intensity as an inverse function of distance from its center. Inspired by \cite{bau2021paint} manipulating $w$ latents only under a user selected mask, we set $s_L = m$ only under the masked regions $M$, while keeping the channels under the non-masked regions $1-M$ unaltered. This gives us a spatially masked style $S$ which specifies a vector for every feature map location:
\begin{align}
    S = S_m \circ M + S_0 \circ (1-M)
  \label{eq:region}
\end{align}
where $S_0$ denotes the original style vector expanded into a tensor, and $S_m$ is the expanded style tensor containing $s_L = m$.
To train $G_c$, we concatenate $G_s(x)$ and the mask $M$ as the input. We keep our modulated ResNet method and also employ reversed training samples for teaching $G_c$ how to turn off lights. Our inputs for discriminative loss terms remains the same as described in  \cref{unsupervised}.

During inference, the user paints over the a photo to create a mask, which is then Gaussian blurred. $G_c$ takes the photo, the mask, and $m$ as input to output a version of that photo with only masked light sources relit to an intensity $m$.

\section{Results}
\begin{table}
  \centering
  \begin{tabular}{@{}l|ccc|c@{}}
     & LPIPS & MSE & RMSE & FID 50k \\
    \toprule
    Ours & 0.205 & 0.084 & 7.54 & 1.842\\
    \midrule 
    No modulation & 0.207 & 0.103 & 8.48 & 1.982 \\
    \midrule
    \shortstack[r]{e4e inversion \cite{tov2021designing}} & 0.498 & 0.115 & 14.35 & 25.79\\
    \shortstack[r]{ADA inversion \cite{karras2020training}} & 0.471 & 0.153 & 13.86 & 19.73*\\

    \bottomrule
  \end{tabular}
  \caption{We compare our method to other candidate relighting methods. We evaluate relighting accuracy using several image similarity metrics against the "ground truth" image from our Lonoff dataset. We also evaluate realism using the Fréchet inception distance (FID) against 50k images from the LSUN Bedrooms dataset. *For FID on the inverter provided by StyleGAN-ADA \cite{karras2020training}, we only evaluate on 2.5k samples due to time limitations. ADA inverts via an optimization loop, which takes significantly longer than a forward pass through an encoder.}
  \label{tab:metrics}
\end{table}
\begin{figure*}
  \centering
    \includegraphics[width=.9\textwidth]{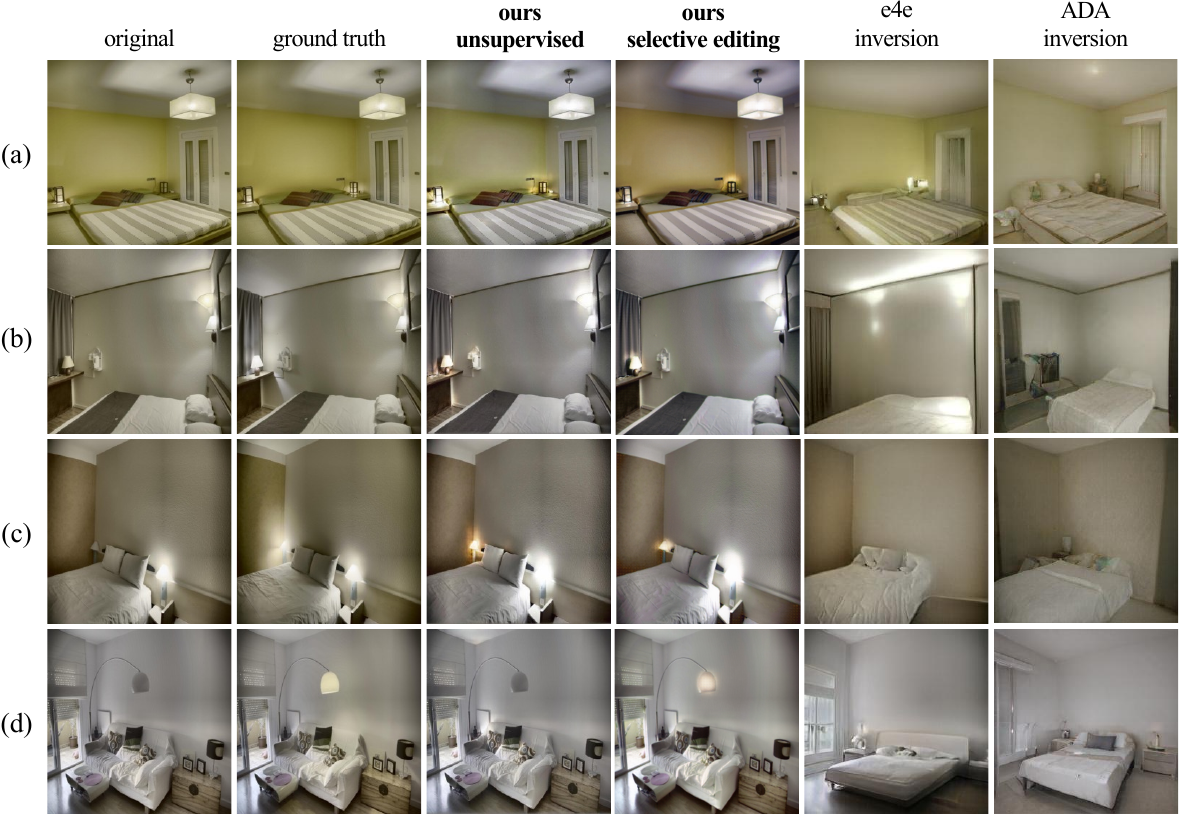}
  \caption{
Qualitative comparison of methods for relighting scenes with visible light sources from the Lonoff dataset. (a), (b), (c) show cases when our unsupervised method is able to recognize and light up unlit light sources. (d) shows a failure case when our unsupervised method is unable to detect the large center lamp, but supervising our selective editing method with a user drawn mask successfully lights up that lamp. We notice that our selective editing method may inaccurately alter the color temperature of the scene as in (a), but can also accurately alter the color temperature as in (d). In comparison, the two inversion methods result in significantly lower quality images that do not  preserve the original image and often times removes the light source. 
  }
  \label{fig:grid_agata_dataset}
\end{figure*}
\begin{figure*}
  \centering
    \includegraphics[width=\textwidth]{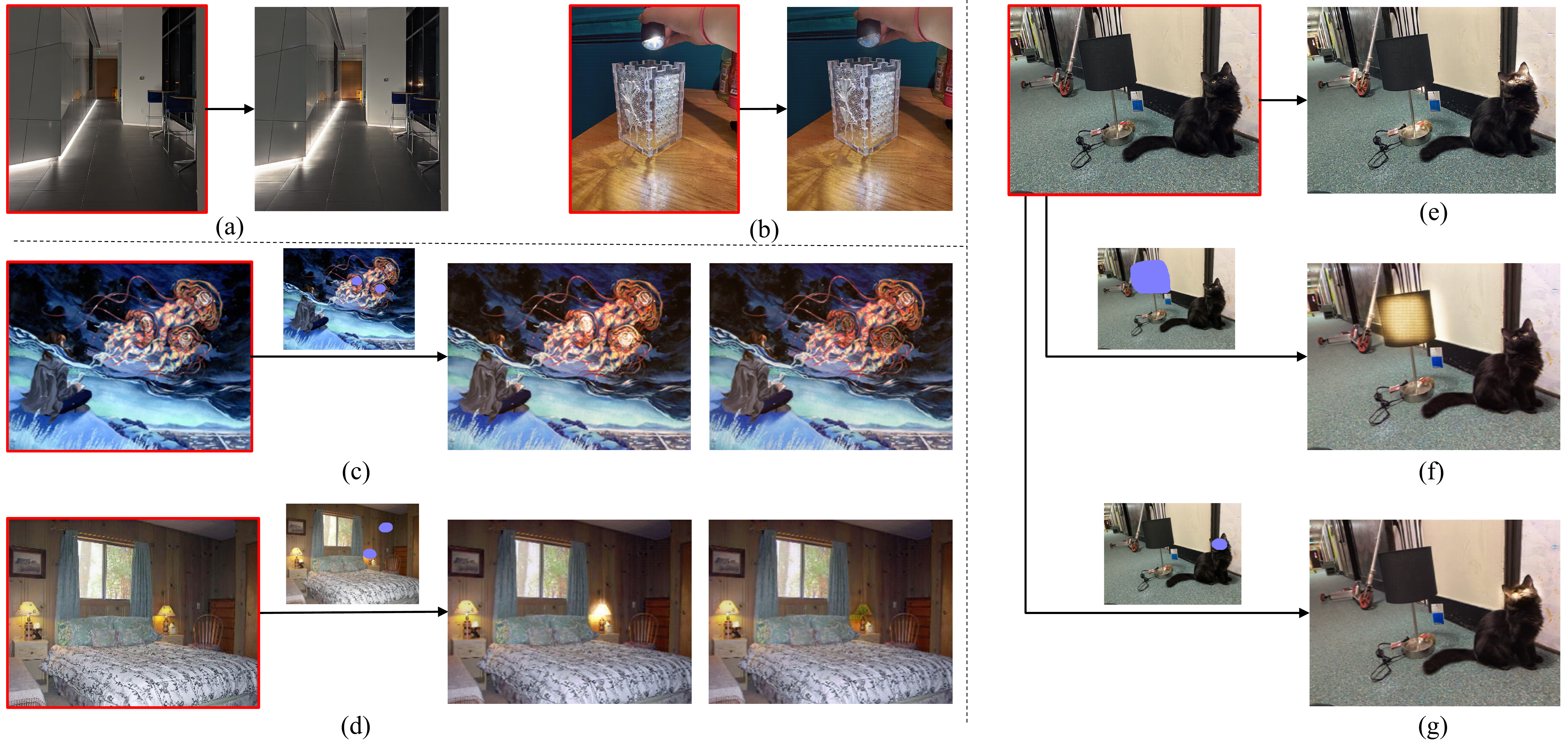}
  \caption{
  We qualitatively examine the capabilities and limitations of our methods on diverse real images containing visible light sources. Images with a red border are the original photos. We discuss each case in Section ~\ref{diverse}.
  }
  \label{fig:examples}
\end{figure*}
\subsection{Comparison of methods on Lonoff}
We use Lonoff to quantitatively evaluate our method against other approaches for local relighting. We focus on the unsupervised task of turning all visible light sources on. We do not quantitatively evaluate on region specific edits because it requires supervision of selecting which lamps to relight. Images with all light sources already lit are set aside as the ``ground truth" for our comparison metrics, while the rest of the images are used as inputs. Images depicting the same scene with a different combination of lights turned on/off would therefore be paired with the same ground truth image. This gives us  $206$ pairs for evaluation. For consistency, we rescale all images to $256 \times 256$ using bilinear interpolation. 

To measure accuracy of relighting in comparison to a ground truth, we use the following quantitative metrics. Mean Squared Error (MSE) measures for pixelwise image similarity, which can be limiting because a small distortion can cause a large pixelwise fluctuation. Grosse et al. \cite{grosse2009ground} propose a more forgiving metric, Local Mean Squared Error (LMSE), which sums the MSE over several windows. We choose for the windows to be $20\times20$ spaced $10$ pixels apart. Lastly, Learned Perceptual Image Patch Similarity (LPIPS)~\cite{zhang2018unreasonable} measures for image perceptual similarity, which is how similar two images are based on human visual perception. For all methods, we translate each input image into three variations of increasing lighting intensity. For the inversion baselines e4e \cite{tov2021designing} and ADA \cite{karras2020training}, we use their method to invert the image into StyleGAN's latent space, identify the style channel that visually best controls for lighting, and activate the channel to three increasing intensities. We use the best out of three relit scenes to calculate each image similarity metric. 

We do not quantitatively evaluate the task of turning lights off because the Lonoff dataset was created such that if all lights were turned off, the scene would be pitch black. 


To evaluate realism, we use the Frechét Inception Distance \cite{heusel2017gans} on $50K$ samples from LSUN Bedrooms against $50K$ relit scenes generated by each method taking randomly sampled LSUN bedrooms as input. We allow each method to relight bedrooms at a randomly chosen intensity (i.e. we choose $m$ randomly out of a range of both positive and negative values). This generates samples of scenes both with lights turned on and off. 

In \cref{tab:metrics}, we see that we outperform other methods for all metrics. It is expected that our main method beats the ablation of not modulating the ResNet blocks. Modulation allows for increased control over lighting intensity. This would generate more diverse lighting conditions that aligns closer with the distribution of real images and more likely relights an image that aligns with the ground truth. Visually, we see in \cref{fig:grid_agata_dataset} that inversion methods result in highly distorted reconstructions that often removes visible light sources from the input. This corroborates with the significantly worse metrics of inversion methods.  

We note that in instances like (d) in  \cref{fig:grid_agata_dataset}, the unsupervised method fails to detect unlit light sources, especially when they deviate from the style of lamps that the pretrained StyleGAN2 generates. In these cases, a user can successfully turn them on using the selective method.

\subsection{Unsupervised and region specific edits on out of domain images} \label{diverse}
In in \cref{fig:teaser} and \cref{fig:examples} we demonstrate our relighting methods on a diverse set of real images that go far beyond the narrow domain of bedrooms that $G_c$ is trained on. While $G_c$ only sees examples of relit lamps, it can adjust the lighting of fire, road signs, and strip lighting. 

In \cref{fig:examples}, (a) and (b) demonstrate our unsupervised method for turning light sources brighter and dimmer, respectively. While (b) successfully turns the flashlight off, it does not remove the reflected light on the table. (c) and (d) show our selective editing method, allowing control over a subset of visible light sources. (c) depicts an artistic painting, showing that our method can work on stylized light sources. (d) masks both a light and a non-light source (a spot on the wall). This is a successful case of controlling a masked light source and accurately preserving the underlying original image under a masked non-light source. (e) demonstrates an interesting failure case of our unsupervised method when a non-light source (the cat) is incorrectly detected. (f) shows that this can be corrected with a user drawn mask on the actual light source. However, (g) shows the cat lighting up after it is masked.

\section{Discussion}
We have introduced the task of local relighting: relighting a scene in which a local light source is visible. By exploiting the ability of a GAN to disentangle factors of variation corresponding to lighting, we have been able to train a model on the challenging task of local relighting without a special training set and without supervision of labels.  We have used the synthesis model to generate an unbounded training set of relit image pairs, which are used to train a pix2pix generative image model.  To facilitate benchmarking of this new task, we have introduced the Lonoff dataset, a new dataset of precisely aligned scene photographs with local lighting changes.  We have found that our method outperforms baseline methods based on GAN inversion, and that our method can be also applied to diverse, out of domain images.  

\section{Ethical Considerations}

While manipulating lighting in an image is an application mainly of interest in artistic and aesthetic applications, we acknowledge that our work could be potentially misused, for example to create realistic manipulated images that misrepresent the state of traffic lights in an evidence photo, or lights in other images relied upon to be realistic. By releasing our code, we hope to enable the community to reproduce our methods and continue to develop countermeasures against misinformation.
\section{Acknowledgements}

We thank Nvidia for publishing weights for pretrained StyleGAN models that make this work possible. We thank Daksha Yadav for her insights, encouragement, and valuable discussions. We are grateful for the support of Signify Lighting Research, DARPA XAI (FA8750-18-C-0004), and the Spanish Ministry of Science, Innovation and Universities (RTI2018-095232-B-C22).

{\small
\bibliographystyle{ieee_fullname}

\begin{thebibliography}{10}\itemsep=-1pt

\bibitem{bau2021paint}
David Bau, Alex Andonian, Audrey Cui, YeonHwan Park, Ali Jahanian, Aude Oliva,
  and Antonio Torralba.
\newblock Paint by word, 2021.

\bibitem{bau2018gan}
David Bau, Jun-Yan Zhu, Hendrik Strobelt, Bolei Zhou, Joshua~B Tenenbaum,
  William~T Freeman, and Antonio Torralba.
\newblock Gan dissection: Visualizing and understanding generative adversarial
  networks.
\newblock {\em arXiv preprint arXiv:1811.10597}, 2018.

\bibitem{el2021ntire}
Majed El~Helou, Ruofan Zhou, Sabine Susstrunk, and Radu Timofte.
\newblock Ntire 2021 depth guided image relighting challenge.
\newblock In {\em Proceedings of the IEEE/CVF Conference on Computer Vision and
  Pattern Recognition}, pages 566--577, 2021.

\bibitem{gafton20202d}
Paul Gafton and Erick Maraz.
\newblock 2d image relighting with image-to-image translation.
\newblock {\em arXiv preprint arXiv:2006.07816}, 2020.

\bibitem{grosse2009ground}
Roger Grosse, Micah~K Johnson, Edward~H Adelson, and William~T Freeman.
\newblock Ground truth dataset and baseline evaluations for intrinsic image
  algorithms.
\newblock In {\em 2009 IEEE 12th International Conference on Computer Vision},
  pages 2335--2342. IEEE, 2009.

\bibitem{gu2020image}
Jinjin Gu, Yujun Shen, and Bolei Zhou.
\newblock Image processing using multi-code gan prior.
\newblock In {\em Proceedings of the IEEE/CVF conference on computer vision and
  pattern recognition}, pages 3012--3021, 2020.

\bibitem{harkonen2020ganspace}
Erik H{\"a}rk{\"o}nen, Aaron Hertzmann, Jaakko Lehtinen, and Sylvain Paris.
\newblock Ganspace: Discovering interpretable gan controls.
\newblock {\em arXiv preprint arXiv:2004.02546}, 2020.

\bibitem{he2016deep}
Kaiming He, Xiangyu Zhang, Shaoqing Ren, and Jian Sun.
\newblock Deep residual learning for image recognition.
\newblock In {\em Proceedings of the IEEE conference on computer vision and
  pattern recognition}, pages 770--778, 2016.

\bibitem{helou2020vidit}
Majed~El Helou, Ruofan Zhou, Johan Barthas, and Sabine S{\"u}sstrunk.
\newblock Vidit: virtual image dataset for illumination transfer.
\newblock {\em arXiv preprint arXiv:2005.05460}, 2020.

\bibitem{heusel2017gans}
Martin Heusel, Hubert Ramsauer, Thomas Unterthiner, Bernhard Nessler, and Sepp
  Hochreiter.
\newblock Gans trained by a two time-scale update rule converge to a local nash
  equilibrium.
\newblock {\em Advances in neural information processing systems}, 30, 2017.

\bibitem{isola2017image}
Phillip Isola, Jun-Yan Zhu, Tinghui Zhou, and Alexei~A Efros.
\newblock Image-to-image translation with conditional adversarial networks.
\newblock In {\em CVPR}, pages 1125--1134, 2017.

\bibitem{jahanian2019steerability}
Ali Jahanian, Lucy Chai, and Phillip Isola.
\newblock On the" steerability" of generative adversarial networks.
\newblock In {\em ICLR}, 2019.

\bibitem{kalantari2016learning}
Nima~Khademi Kalantari, Ting-Chun Wang, and Ravi Ramamoorthi.
\newblock Learning-based view synthesis for light field cameras.
\newblock {\em ACM Transactions on Graphics (TOG)}, 35(6):1--10, 2016.

\bibitem{karras2020training}
Tero Karras, Miika Aittala, Janne Hellsten, Samuli Laine, Jaakko Lehtinen, and
  Timo Aila.
\newblock Training generative adversarial networks with limited data.
\newblock {\em arXiv preprint arXiv:2006.06676}, 2020.

\bibitem{karras2019style}
Tero Karras, Samuli Laine, and Timo Aila.
\newblock A style-based generator architecture for generative adversarial
  networks.
\newblock In {\em Proceedings of the IEEE/CVF Conference on Computer Vision and
  Pattern Recognition}, pages 4401--4410, 2019.

\bibitem{karras2020analyzing}
Tero Karras, Samuli Laine, Miika Aittala, Janne Hellsten, Jaakko Lehtinen, and
  Timo Aila.
\newblock Analyzing and improving the image quality of stylegan.
\newblock In {\em Proceedings of the IEEE/CVF Conference on Computer Vision and
  Pattern Recognition}, pages 8110--8119, 2020.

\bibitem{mahajan2007theory}
Dhruv Mahajan, Ira~Kemelmacher Shlizerman, Ravi Ramamoorthi, and Peter
  Belhumeur.
\newblock A theory of locally low dimensional light transport.
\newblock {\em ACM SIGGRAPH 2007 papers}, pages 62--es, 2007.

\bibitem{malzbender2001polynomial}
Tom Malzbender, Dan Gelb, and Hans Wolters.
\newblock Polynomial texture maps.
\newblock In {\em Proceedings of the 28th annual conference on Computer
  graphics and interactive techniques}, pages 519--528, 2001.

\bibitem{shen2020interpreting}
Yujun Shen, Jinjin Gu, Xiaoou Tang, and Bolei Zhou.
\newblock Interpreting the latent space of gans for semantic face editing.
\newblock In {\em Proceedings of the IEEE/CVF Conference on Computer Vision and
  Pattern Recognition}, pages 9243--9252, 2020.

\bibitem{sloan2003clustered}
Peter-Pike Sloan, Jesse Hall, John Hart, and John Snyder.
\newblock Clustered principal components for precomputed radiance transfer.
\newblock {\em ACM Transactions on Graphics (TOG)}, 22(3):382--391, 2003.

\bibitem{tov2021designing}
Omer Tov, Yuval Alaluf, Yotam Nitzan, Or Patashnik, and Daniel Cohen-Or.
\newblock Designing an encoder for stylegan image manipulation.
\newblock {\em ACM Transactions on Graphics (TOG)}, 40(4):1--14, 2021.

\bibitem{Viazovetskyi2020StyleGAN2DF}
Yuri Viazovetskyi, Vladimir Ivashkin, and Evgenii Kashin.
\newblock Stylegan2 distillation for feed-forward image manipulation.
\newblock {\em European Conference on Computer Vision}, abs/2003.03581, 2020.

\bibitem{wang2018high}
Ting-Chun Wang, Ming-Yu Liu, Jun-Yan Zhu, Andrew Tao, Jan Kautz, and Bryan
  Catanzaro.
\newblock High-resolution image synthesis and semantic manipulation with
  conditional gans.
\newblock In {\em Proceedings of the IEEE conference on computer vision and
  pattern recognition}, pages 8798--8807, 2018.

\bibitem{wang2021multi}
Yuanzhi Wang, Tao Lu, Yanduo Zhang, and Yuntao Wu.
\newblock Multi-scale self-calibrated network for image light source transfer.
\newblock In {\em Proceedings of the IEEE/CVF Conference on Computer Vision and
  Pattern Recognition}, pages 252--259, 2021.

\bibitem{wu2021stylespace}
Zongze Wu, Dani Lischinski, and Eli Shechtman.
\newblock Stylespace analysis: Disentangled controls for stylegan image
  generation.
\newblock In {\em Proceedings of the IEEE/CVF Conference on Computer Vision and
  Pattern Recognition}, pages 12863--12872, 2021.

\bibitem{xiao2018unified}
Tete Xiao, Yingcheng Liu, Bolei Zhou, Yuning Jiang, and Jian Sun.
\newblock Unified perceptual parsing for scene understanding.
\newblock In {\em Proceedings of the European Conference on Computer Vision
  (ECCV)}, pages 418--434, 2018.

\bibitem{xu2018deep}
Zexiang Xu, Kalyan Sunkavalli, Sunil Hadap, and Ravi Ramamoorthi.
\newblock Deep image-based relighting from optimal sparse samples.
\newblock {\em ACM Transactions on Graphics (ToG)}, 37(4):1--13, 2018.

\bibitem{yang2021multi}
Hao-Hsiang Yang, Wei-Ting Chen, Hao-Lun Luo, and Sy-Yen Kuo.
\newblock Multi-modal bifurcated network for depth guided image relighting.
\newblock In {\em Proceedings of the IEEE/CVF Conference on Computer Vision and
  Pattern Recognition}, pages 260--267, 2021.

\bibitem{yazdani2021physically}
Amirsaeed Yazdani, Tiantong Guo, and Vishal Monga.
\newblock Physically inspired dense fusion networks for relighting.
\newblock In {\em Proceedings of the IEEE/CVF Conference on Computer Vision and
  Pattern Recognition}, pages 497--506, 2021.

\bibitem{zhang2018unreasonable}
Richard Zhang, Phillip Isola, Alexei~A Efros, Eli Shechtman, and Oliver Wang.
\newblock The unreasonable effectiveness of deep features as a perceptual
  metric.
\newblock In {\em Proceedings of the IEEE conference on computer vision and
  pattern recognition}, pages 586--595, 2018.

\bibitem{Zhang2021DatasetGANEL}
Yuxuan Zhang, Huan Ling, Jun Gao, K. Yin, Jean-Francois Lafleche, Adela
  Barriuso, Antonio Torralba, and Sanja Fidler.
\newblock Datasetgan: Efficient labeled data factory with minimal human effort.
\newblock {\em 2021 IEEE/CVF Conference on Computer Vision and Pattern
  Recognition (CVPR)}, pages 10140--10150, 2021.

\bibitem{zhou2017scene}
Bolei Zhou, Hang Zhao, Xavier Puig, Sanja Fidler, Adela Barriuso, and Antonio
  Torralba.
\newblock Scene parsing through ade20k dataset.
\newblock In {\em Proceedings of the IEEE conference on computer vision and
  pattern recognition}, pages 633--641, 2017.

\bibitem{zhu2020domain}
Jiapeng Zhu, Yujun Shen, Deli Zhao, and Bolei Zhou.
\newblock In-domain gan inversion for real image editing.
\newblock In {\em European conference on computer vision}, pages 592--608.
  Springer, 2020.

\end{thebibliography}

}

\newpage
\onecolumn
\noindent\title{\Large{\textbf{Appendices}}}
\begin{appendix}

\section{Identification of the lighting channel $s_L$}\label{sec:channel_id}
Our method for identifying the lighting channel $s_L$ introduces a new approach to controlling a large image generation network through one annotated example. Our work is inspired by \cite{wu2021stylespace}, which used a pretrained classifier or between 10-30 example images to identify StyleSpace channels for controlling specific scene attributes, and by \cite{Zhang2021DatasetGANEL}, which used between 16-40 manually segmented images to train a small decoder to segment any image generated by a GAN. Unlike \cite{wu2021stylespace} and \cite{Zhang2021DatasetGANEL}, we observe that using just a single manually annotated image can be sufficient. 

We begin by generating 12 StyleGAN bedroom scenes and manually selecting a single generated image $x = G_s(z)$ that depicts a lamp casting light that is reflected on a wall. On $x$, we manually select the region occupied by the reflected light as the target control area $T$. We are interested in the network's ability to model the propagation of light throughout the scene, rather than to merely mimic the shape of the light fixture itself. We then iterate through the 5120 StyleSpace channels $s$ that control feature maps in this version of StyleGAN. For each $s$, we set $s=0$ to generate a modified image $x^{\prime} = G_s^{\prime}(z)$. Each $s$ is ranked by $\sum |x-x^{\prime}| \circ T$, which is the pixelwise sum of the absolute difference that lands in the target control area. $s_L$ is selected as the single highest ranking channel. 
\begin{figure*}[!htb]
  \centering
    \includegraphics[width=1\textwidth]{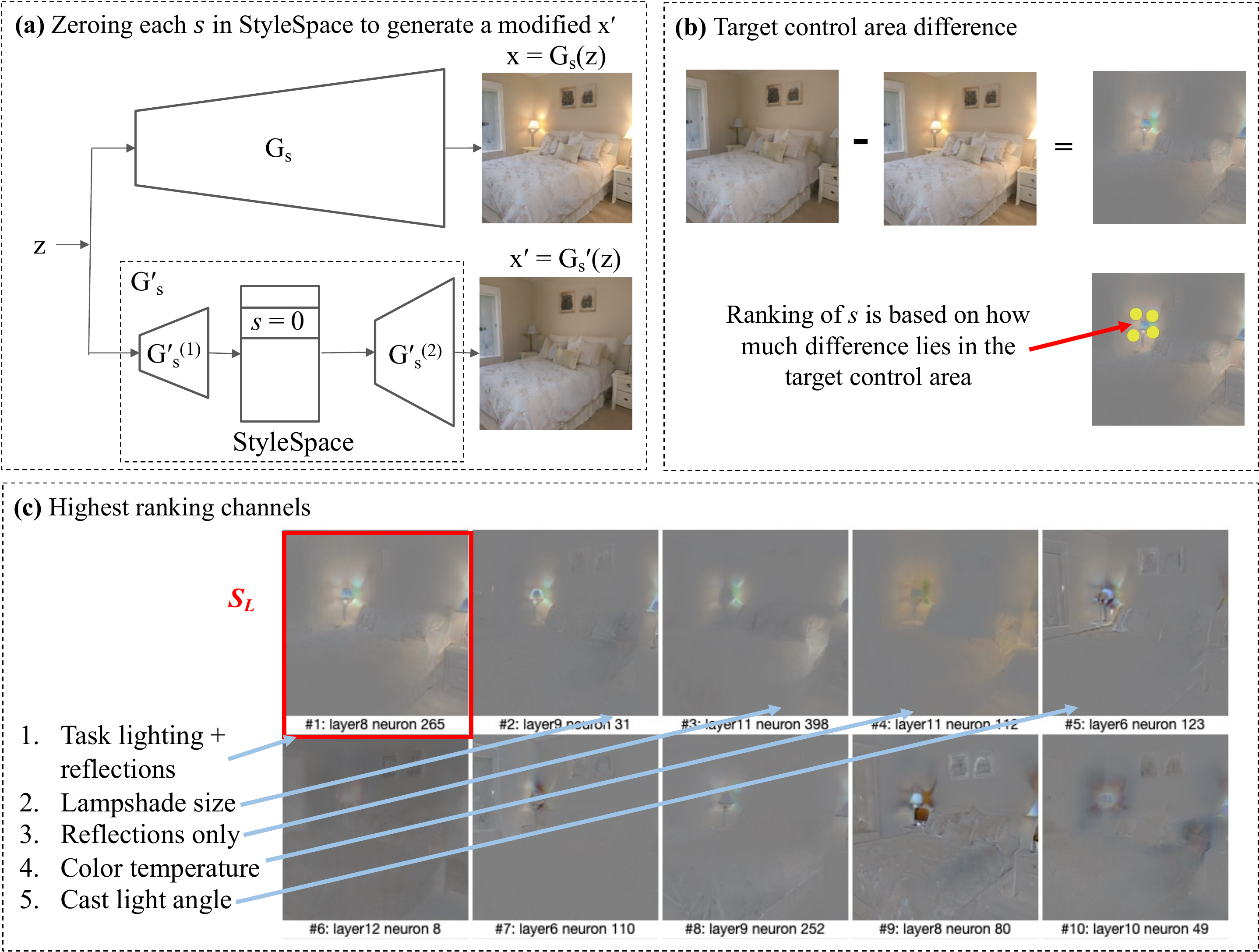}
  \caption{
  (a) We iterate through and zero out each channel $s$ to generate modified images $x^{\prime}$. (b) Each $s$ is then ranked based on the difference in the manually chosen target control area. (c) shows the highest ranking channels. Notice the five highest ranking channels control for slightly difference characteristics of lighting, such as reflections, lampshade size, color temperature, and light angle. The single highest ranking channel modifies lighting in the way that best suits our objective of predicting light propagation in a scene, so we select it as $s_L$. 
  }
  \label{fig:channelidfig}
\end{figure*}
\clearpage

\section{Modulation of Multiple Channels}\label{sec:multichannel}
Here we show that our work extends the idea of channel modulation beyond the StyleSpace channel for lamp lighting. We apply the technique described in Appendix \ref{sec:channel_id} for finding a channel that controls window light intensity. By modifying the two channels for lamps and windows respectively, we can generate a dataset of paired samples containing variations in both lamp and window lighting.

\begin{figure*}[!htb]
  \centering
    \includegraphics[width=1\textwidth]{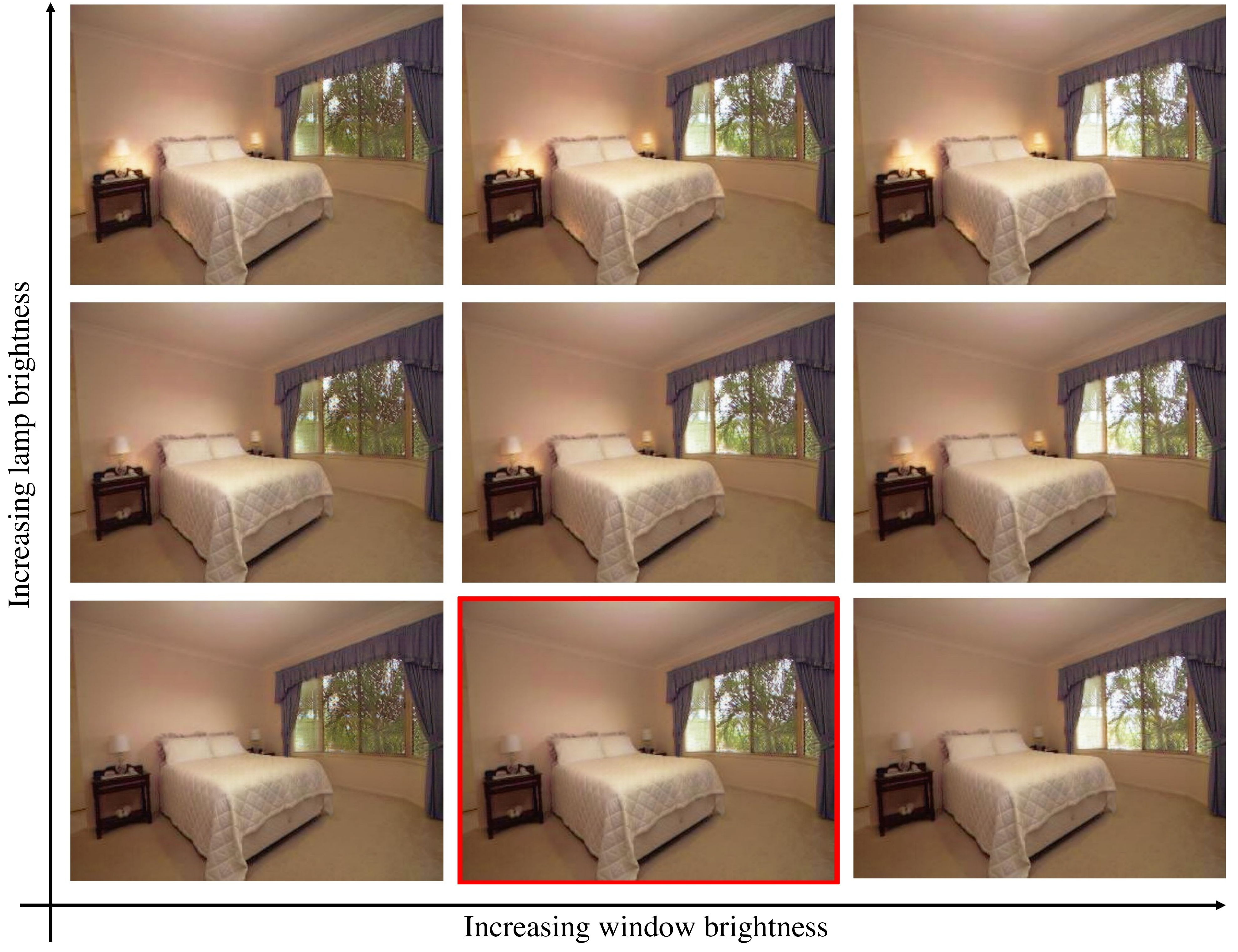}
  \caption{
Lamp and window modulation example on a real image. The original image is boxed in red. We show that during inference, lamp and window lighting can be controlled separately by inputting different modulation scalars. No further supervision is required to separately control different light source types. 
  }
  \label{fig:lamp+window_mods}
\end{figure*}

We can then use the dataset to train a pix2pix with its ResNet bottleneck modulated by two scalars instead of just a single value. We find that pix2pix learns to control both disentangled scalars, allowing us to control lamps and windows separately. Some qualitative results are illustrated in Fig.~\ref{fig:lamp+window_mods}. 

\clearpage

\section{Additional Qualitative Examples}\label{sec:qualitative_exmaples}

\begin{figure*}[!htb]
\centering%
\includegraphics[width=1\textwidth,trim=0 0 0 0]{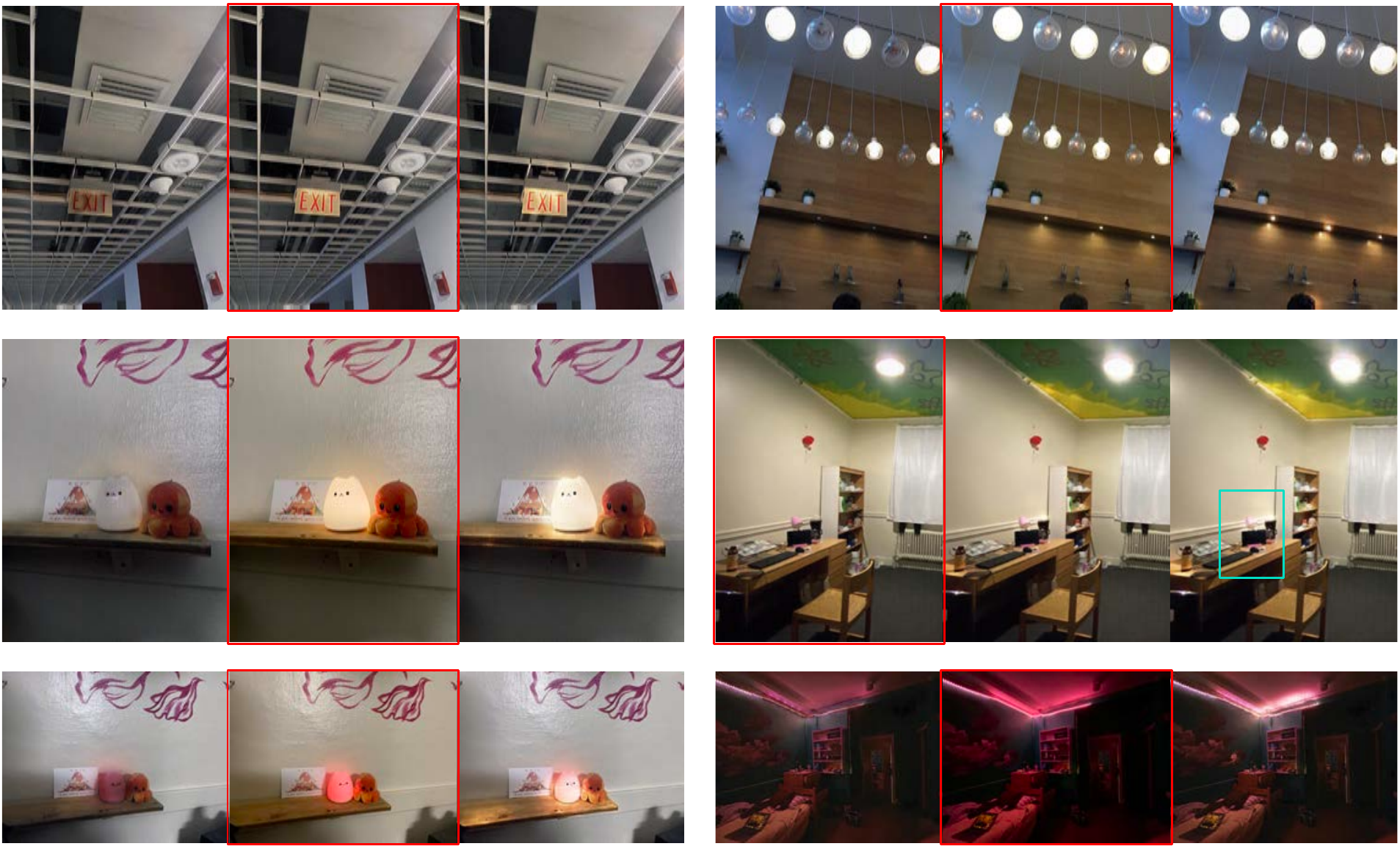}\\
\captionof{figure}{More examples of our unsupervised method on diverse images. In each panel, the red box shows the original image, and lower and/or higher light intensities are illustrated in its left and right images, respectively. The example in the middle row of the right column shows an originally unlit pink desk lamp (boxed in blue) being turned on by our method. }\label{fig:moreunsupervised-crop}
\end{figure*}

Additional qualitative examples of our unsupervised relighting method are shown in Fig.~\ref{fig:moreunsupervised-crop}. Note these images are quite outside of the distribution of the bedroom dataset that we generated from StyleGAN2 for training pix2pixHD.

\begin{figure*}[!htb]
  \centering
    \includegraphics[width=1\textwidth]{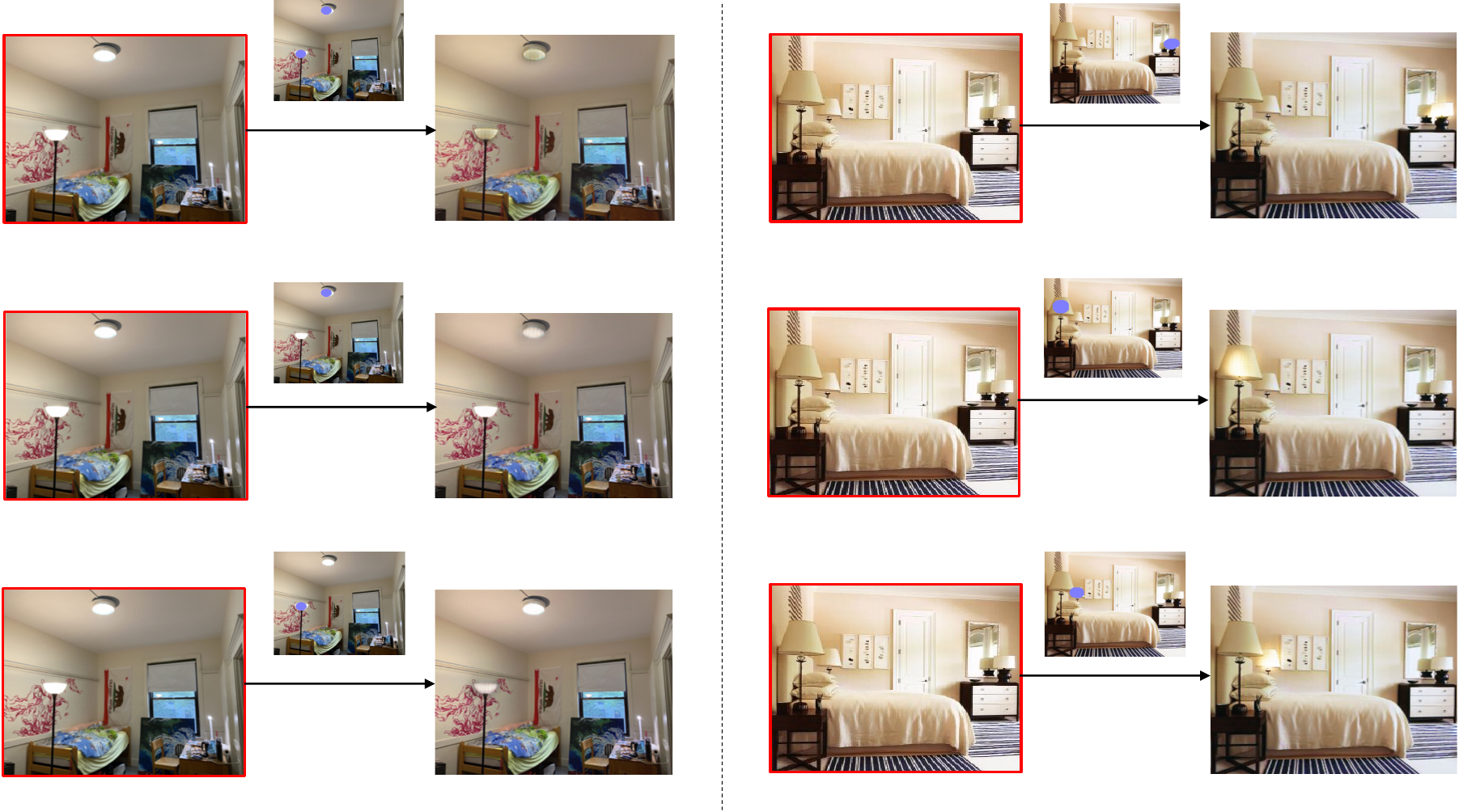}
  \caption{
Demonstrations of our user selective method. Examples on the left demonstrate different combinations of lights being turned off, and examples on the right demonstrate different lights being turned on.
  }
  \label{fig:turn_off}
\end{figure*}

Figure~\ref{fig:turn_off} illustrates more examples of our ``user selective'' editing, where the user can choose which light(s) to edit. For ease of visualization and comparison, we show the edits on the same bedroom image. 

\clearpage
\

\section{Lonoff Dataset Samples}\label{sec:Lonoff_samples}
\begin{figure*}[!htb]
\centering%
\includegraphics[width=1\textwidth,trim=0 0 0 0]{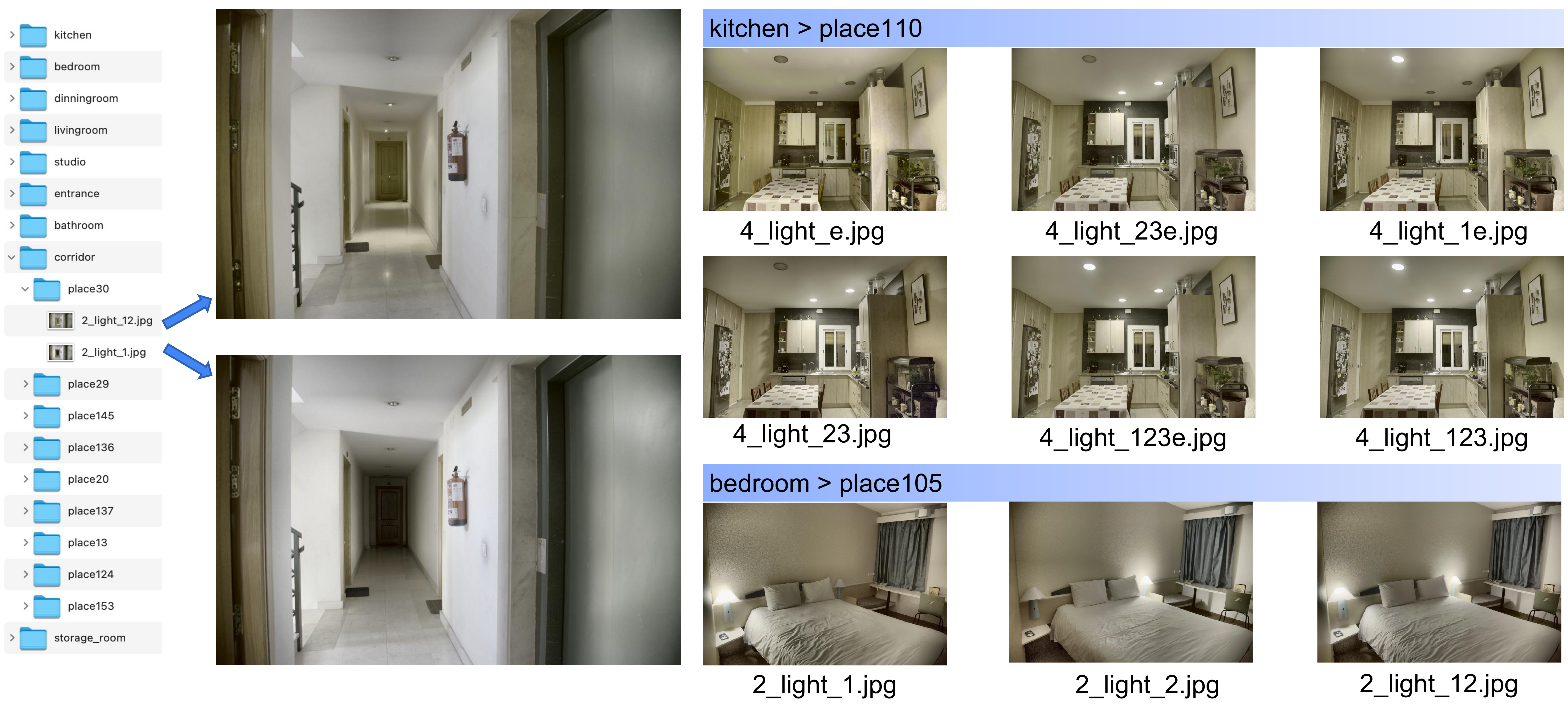}\\
\captionof{figure}{A snapshot of our dataset ``Lonoff'' with samples in three categories. }\label{fig:Lonoff_samples}
\end{figure*}

Figure~\ref{fig:Lonoff_samples} illustrates a snapshot of the organization of Lonoff along with some examples. Each image's filename contains its light information. For instance, in the ``kitchen'' category, the ``place110'' directory contains images of a kitchen with $4$ light sources. The subsequent letter ``e'' corresponds to ``external light,'' and the subsequent numbers (ex. $23$) correspond to the indices of light sources, scanning from left to right. 

\end{appendix}

\end{document}